\documentclass[10pt,twocolumn,letterpaper]{article}
\usepackage[pagebackref=true,breaklinks=true,letterpaper=true,colorlinks,bookmarks=false]{hyperref}
\usepackage{cvpr}
\usepackage{times}
\usepackage{epsfig}
\usepackage{graphicx}
\usepackage{amsmath}
\usepackage{amssymb}

\pdfoutput=1
\cvprfinalcopy 

\def\cvprPaperID{****} 

\begin{document}

\title{LBVCNN: Local Binary Volume Convolutional Neural Network for Facial Expression Recognition from Image Sequences}

\author{Sudhakar Kumawat$^{1}$ \quad\quad Manisha Verma$^{2}$ \quad\quad Shanmuganathan Raman$^{1}$\\
$^{1}$Indian Institute of Technology Gandhinagar, India, \hspace{0.1em} $^{2}$Osaka University, Japan\\
$^{1}$\small\{\tt sudhakar.kumawat, shanmuga\}@iitgn.ac.in, $^{2}${\tt\small manisha.verma89}@gmail.com} 

\maketitle

\begin{abstract}
	Recognizing facial expressions is one of the central problems in computer vision. Temporal image sequences have useful spatio-temporal features for recognizing expressions. 
	In this paper, we propose a new 3D Convolution Neural Network (CNN) that can be trained end-to-end for facial expression recognition on temporal image sequences without using facial landmarks. More specifically, a novel 3D convolutional layer that we call Local Binary Volume (LBV) layer is proposed. The LBV layer, when used with our newly proposed LBVCNN network, achieve comparable results compared to state-of-the-art landmark-based or without landmark-based models 
	on image sequences from CK+, Oulu-CASIA, and UNBC McMaster shoulder pain datasets. Furthermore,  our LBV layer reduces the number of trainable parameters by a significant amount when compared to a conventional 3D convolutional layer. As a matter of fact, when compared to a $3\times3\times3$ conventional 3D convolutional layer, the LBV layer uses 27 times less trainable parameters.
\end{abstract}


\section{Introduction}
Facial expressions are subtle signals of a larger communication process.
They express one's feelings in the form of facial muscle displacements. A simple smile can indicate our liking, while a frown might show our displeasure. Thus, understanding facial expressions is an important part of our communication.
In computer vision, facial expression recognition deals with the problem of recognizing basic human expressions from video or image data. The problem has many applications in the field of  computer science, medicine, psychology, and other related areas. 
\begin{figure}[t]
	\begin{center}
		\includegraphics[width=.14\textwidth]{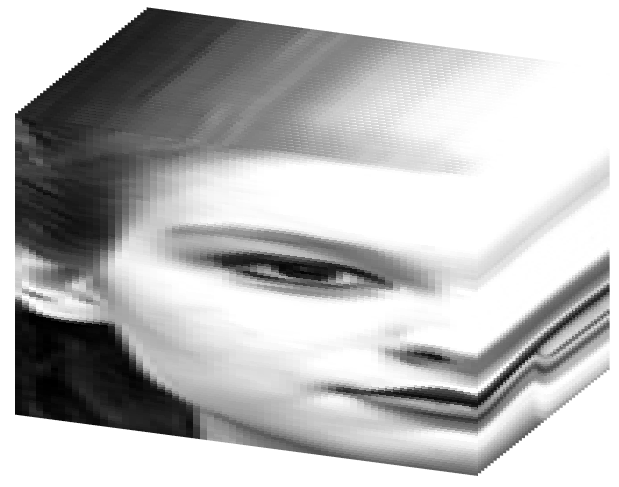} 
		\includegraphics[width=.14\textwidth]{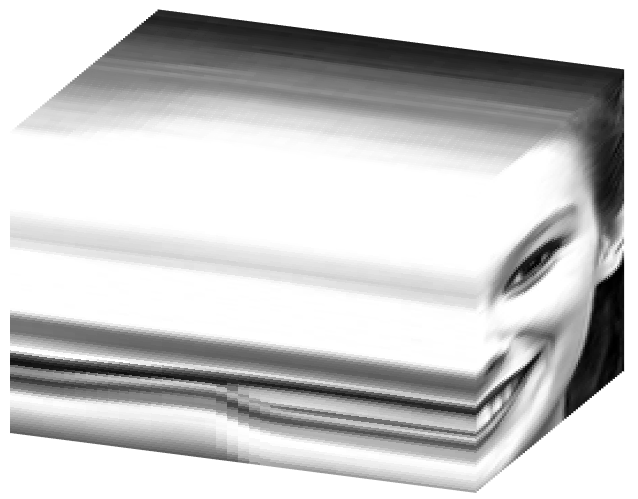}
		\includegraphics[width=.14\textwidth]{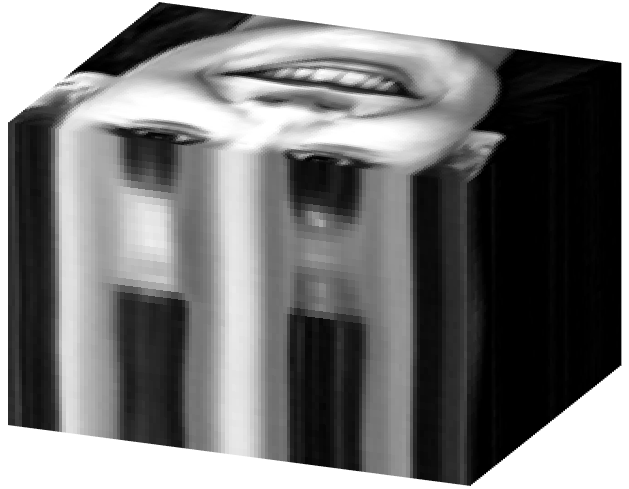}\\
		\includegraphics[width=.14\textwidth]{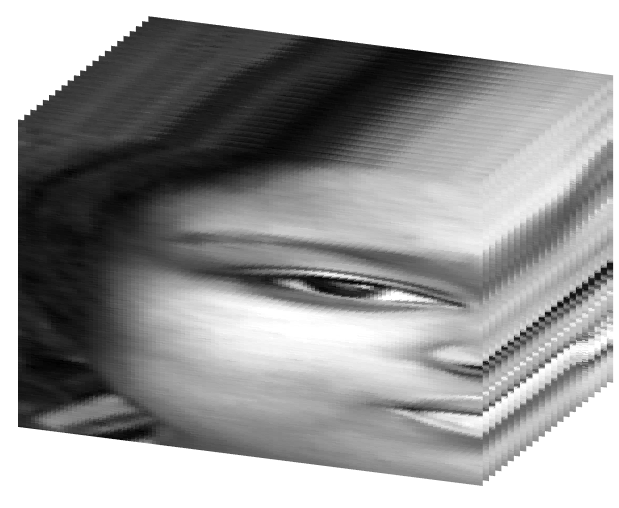} 
		\includegraphics[width=.14\textwidth]{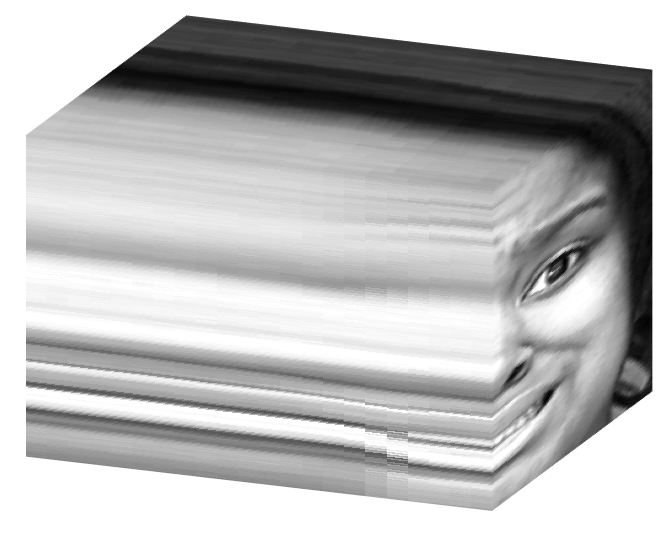}
		\includegraphics[width=.14\textwidth]{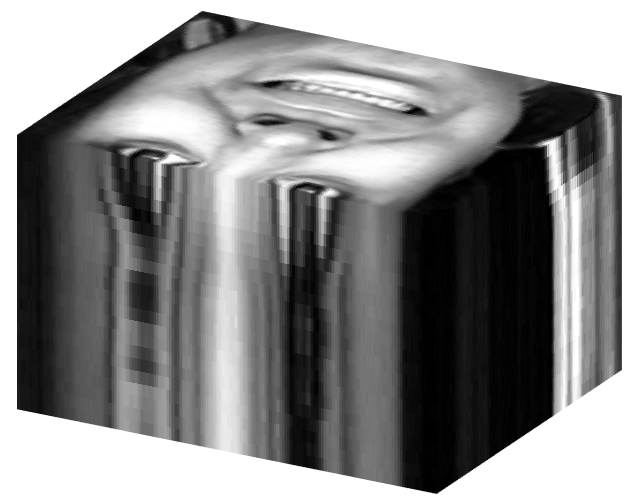}
		\caption{Space-time transitions in the third dimension for XY, YT, and XT spaces.}	
		\label{fig:3dexample}
	\end{center}
\end{figure}
Part of the research on this problem is focused on recognizing facial expressions from static images \cite{ptucha2011manifold,mollahosseini2016going,sariyanidi2017learning,meng2017identity,li2017reliable,fan2016video,dhall2015video,cai2017island,zhao2016peak}. Although this approach is effective in extracting spatial information, it fails to capture morphological and contextual variations of the expression process. Recent methods aim to solve this problem by using temporal image sequences and utilize both spatial and temporal variations to give better recognition systems \cite{jeni2014spatio,liu2014learning,sikka2012exploring,sanin2013spatio,wang2013capturing}. Very recent methods use geometric features like temporal variations in facial landmarks along with temporal image sequences to achieve state-of-the-art results \cite{ghimire2013geometric,youssif2011automatic,jung2015joint,zhang2017facial}. Facial landmarks boost the accuracy of models by supplying discriminant information that steer the expression recognition process, especially with deep learning. However, detecting accurate facial landmarks is a problem by itself. Difficult visual conditions like illumination, resolution, and alignment may further make facial landmarks detection difficult. Recently, Steger \emph{et al.} studied the effects of trivial image distortions  like rotation and Gaussian noise on the performance of facial landmarks detection algorithms \cite{steger2016failure}. The study, which is a first of its kind, showed that even state-of-the-art facial landmarks detection models like Uricar \cite{uvrivcavr2015real} and Kazemi \cite{kazemi2014one} are vulnerable to image distortions. This emphasizes the need for a method that can be used in the domain of facial applications like facial expression recognition and has an accurate performance at par with the state-of-the-art methods, while not using facial landmarks.  \\


In this paper, we propose a simple deep 3D Convolutional Neural Network (CNN) that can be trained end-to-end on temporal image sequences without using any extra information like facial landmarks. Our work is inspired from Volume Local Binary Patterns (VLBP) \cite{zhao2007dynamic} and recently proposed Local Binary Convolutional Neural Network (LBCNN) \cite{juefei2017local}. VLBP takes a 3D neighborhood of each pixel of every frame of a video and generates the corresponding 3D LBP. LBCNN replaces the conventional 2D convolutional layer of CNN by a Local Binary Convolutional (LBC) layer that exploits LBP concept in a CNN architecture. Normally, a video sequence is understood as a stack of XY planes along T axis, but it is easy to see that it can also be seen as a stack of XT planes along Y axis and YT planes along X axis, respectively. The XT and YT planes too have information about the space-time transitions as shown in Fig.~\ref{fig:3dexample}. Our proposed network that we call Local Binary Volume Convolutional Neural Network (LBVCNN) captures these transitions by using three small networks LBVCNN-XY, LBVCNN-XT, and LBVCNN-YT. Each of these small networks consists of our newly proposed Local Binary Volume (LBV) layer which is a 3D variant of the Local Binary Convolution (LBC) layer of the LBCNN network. The three 3D convolutional neural networks LBVCNN-XY, LBVCNN-XT, and LBVCNN-YT are trained on the three orthogonal sides XY, XT, and YT respectively of a video cuboid. Finally, these fully trained networks are combined and then fine-tuned. The main motivation behind this idea is that the local texture information is significant in spatial structure (facial texture) as well as in minor spatio-temporal fluctuations (see Fig.~\ref{fig:3dexample}).\\ 

The main contributions of this paper are summarized as follows.

\begin{itemize}
	\item We propose a new network  called Local Binary Volume Convolutional Neural Network (LBVCNN) that can be trained end-to-end on facial expression image sequences without using landmarks.
	\item Our network uses significantly fewer parameters and has a lower computational cost when compared to the other conventional 3D CNN networks.
	\item We have validated the proposed method on CK+, Oulu-CASIA, and UNBC McMaster dataset.   
\end{itemize}

The rest of the paper is organized as follows. Section~\ref{sect:rw} provides an overview of the relevant works. Section~\ref{sect:lbvcnn} discusses
the architecture of our proposed network LBVCNN. Section~\ref{sect:exp} discusses the datasets used for
the experiments along with the training and the implementation details. It also discusses comparison with
the state-of-the-art approaches. Section~\ref{sect:conclusion} provides the conclusion and  the future work that can be performed.

\section{Related Work}\label{sect:rw}

\begin{figure*}[t]
	\centering
	\includegraphics[width=\textwidth]{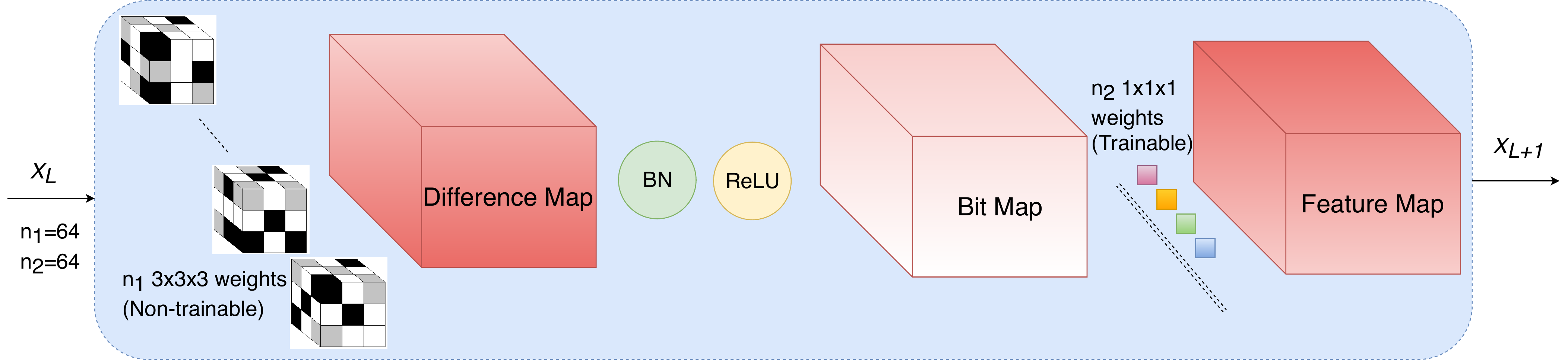}
	\caption{The proposed Local Binary Volume (LBV) block. BN- Batch Normalization. ReLU- Rectified Linear Unit}
	\label{fig:lbcnn}
\end{figure*}

Many existing techniques target facial expression recognition in images and video sequences \cite{sandbach2012static}. Earlier works on facial expression  recognition were concentrated on images \cite{ptucha2011manifold,mollahosseini2016going,sariyanidi2017learning,meng2017identity,li2017reliable,fan2016video,dhall2015video,cai2017island,zhao2016peak}. However, they do not consider temporal variations. Facial expression process is a dynamic event which takes minute motion changes through time into account. Before the era of deep learning, hand-crafted features were used to extract spatio-temporal information and to classify facial expressions. We give a brief overview of various methods that have achieved good performance on facial expression video sequences below.  \\   

\noindent\textbf{Hand-Crafted Feature-Based Methods.} For facial expression analysis in video sequences, many image-based features are extended in order to get temporal features along with spatial information such as LBP-TOP \cite{zhao2007dynamic}, 3D-HOG \cite{klaser2008spatio}, and 3D-SIFT \cite{scovanner20073}. Jain \emph{et al.} used conditional random fields and manually created shape-appearance features for temporal modeling of each facial shape \cite{jain2011facial}. Sanin \emph{et al.} proposed spatio-temporal covariance descriptors using Riemannian locality preserving projection approach for action and gesture recognition \cite{sanin2013spatio}. Wang \emph{et al.} proposed an Interval Temporal Bayesian Network (ITBN) for capturing complex spatio-temporal relations among facial muscles \cite{wang2013capturing}. Liu \emph{et al.} proposed an expressionlet-based spatio-temporal manifold method for dynamic expression recognition \cite{liu2014learning}. Ptucha \emph{et al.} proposed a Manifold-based Sparse Representation (MSR) for expression recognition by mapping features in low dimensional manifolds using supervised locality preserving projections \cite{ptucha2011manifold}. Recently, Sikka \emph{et al.} proposed a Latent Ordinal Model (LOMo) for facial expression recognition in videos \cite{sikka2016lomo}. LOMo integrates features extracted from SIFT around the facial landmarks and LBP using a weakly supervised classifier to learn the expressions as hidden variables.\\

\noindent\textbf{Deep Learning-Based Methods.} Deep learning-based models have achieved state-of-the-art results in facial expression recognition. 
Liu \emph{et al.} applied 3D CNN  with deformable action part constraints (3D CNNDAP) to the problem of expression recognition \cite{liu2014deeply}. 
Recent models use geometric features like facial landmarks to further boost the accuracy. Jung \emph{et al.} proposed two separate networks called DTAN and DTGN and jointly fine-tuned the two networks to achieve state-of-the-art performance \cite{jung2015joint}. The DTAN network is a simple 3D convolutional network that captures spatio-temporal information from temporal image sequences. The DTGN network is a fully-connected network that captures temporal variations in facial landmarks. Guo \emph{et al.} improved Jung \emph{et al.}'s result and trained a spatial network (MSCNN) and a temporal network (PHRNN) separately and jointly fine-tuned them \cite{zhang2017facial}. MSCNN is a simple convolutional network on peak expression images. PHRNN is a collection of subnets (recurrent neural networks) that are connected in a binary tree-like structure. Facial landmarks are divided into four parts and passed at the bottom of this structure and the outputs of the subnets are concatenated at the next layer. The process is repeated for upper layers and the final layer is a softmax classification layer.

\section{Our Approach} \label{sect:lbvcnn}


We propose a 3D convolutional neural network based architecture. Our idea is inspired from Volume Local Binary Pattern (VLBP) \cite{zhao2007dynamic} and the recently proposed Local Binary Convolutional Neural Network (LBCNN) \cite{juefei2017local}.
We give a brief description of the works that inspired our model below. A detailed description of our network is given in Section~\ref{sec:LBVCNN}. \\

\noindent\textbf{Local Binary Pattern.} Local Binary Pattern (LBP) was proposed by Ojala \emph{et al.} \cite{ojala1996comparative}. It computes a binary pattern using each pixel of an image. Every pixel of the image is treated as a center pixel and thresholded with neighborhood pixels. It assigns 0 or 1 to a neighborhood pixel if it is lesser or greater than the center pixel, respectively. Illumination invariance is an important property which makes LBP robust and it has been used in many computer vision problems for feature extraction. For a center pixel $I_c$ and a neighboring pixel $I_i$ ($i=1,2,..,p$), LBP can be formalized as follows.
\begin{equation}
\begin{aligned}&\text{LBP}_{p,r} =\sum \limits _{i=1}^{p}F (I_i -I_c) \times {2^{i-1}}\end{aligned} 
\label{eq2}
\end{equation}
\begin{equation}
\hspace{.1in} \begin{aligned}&F (I)=\left\{ {\begin{array}{l@{\quad }l} 1,&I\ge 0. \\ 0,&\text{ otherwise.} \\ \end{array}} \right. \end{aligned}
\end{equation}
\noindent Here, $p$ and $r$ are the number of neighboring pixels and the radius, respectively. After construction of local binary pattern map, a histogram is created to form the feature descriptor which can be used for classification. \\

\noindent\textbf{Volume Local Binary Pattern.} In order to make LBP useful for dynamic video sequences, Zhao and Pietikainen proposed Volume LBP (VLBP) for dynamic texture recognition \cite{zhao2007dynamic}. VLBP takes a 3D neighborhood of each pixel of every frame and generates the corresponding 3D LBP. To make it computationally simple, LBP is extracted from three orthogonal planes (XY, XT \& YT) corresponding to a center pixel and called as LBP-TOP (Local Binary Pattern - Three Orthogonal Planes). Finally, all the three LBP histograms are concatenated in order to form a feature descriptor which can be fed into a classification algorithm. The feature descriptor combines motion features with spatial features and extracts significant information from the video sequences. Note that, although LBP-TOP is computationally cheap, it is not equivalent to VLBP \cite{zhao2007dynamic}. This is because, it does not take into account all the pixels in the 3D neighborhood of a center pixel as done by VLBP. In LBP-TOP, only the co-occurrences of the local binary patterns on three orthogonal planes are taken into account \cite{zhao2007dynamic}. \\

\noindent\textbf{Local Binary Convolutional Neural Network.} Xu \emph{et al.} \cite{juefei2017local} proposed Local Binary Convolutional Neural Network (LBCNN). In this network, the conventional convolutional layer of CNN is replaced by a Local Binary Convolutional (LBC) layer which is a generalized version of simple LBP. The LBC layer broadly consists of two sub-layers. The first layer involves convolving the input with fixed non-trainable filters of size $3\times3$ in order to get a difference map, followed by a ReLU activation to get an approximate local binary bit-map. The non-trainable filters contain values sampled from the set \{-1,0,1\} using Bernoulli distribution. The second layer is trainable and involves 1$\times $1 convolutions on the output of the first layer in order to get feature maps. This architecture significantly reduces the number of trainable parameters as it involves training of only  $1\times 1$ filters.


\begin{figure*}[t]
	\centering
	\includegraphics[width=1.0\textwidth]{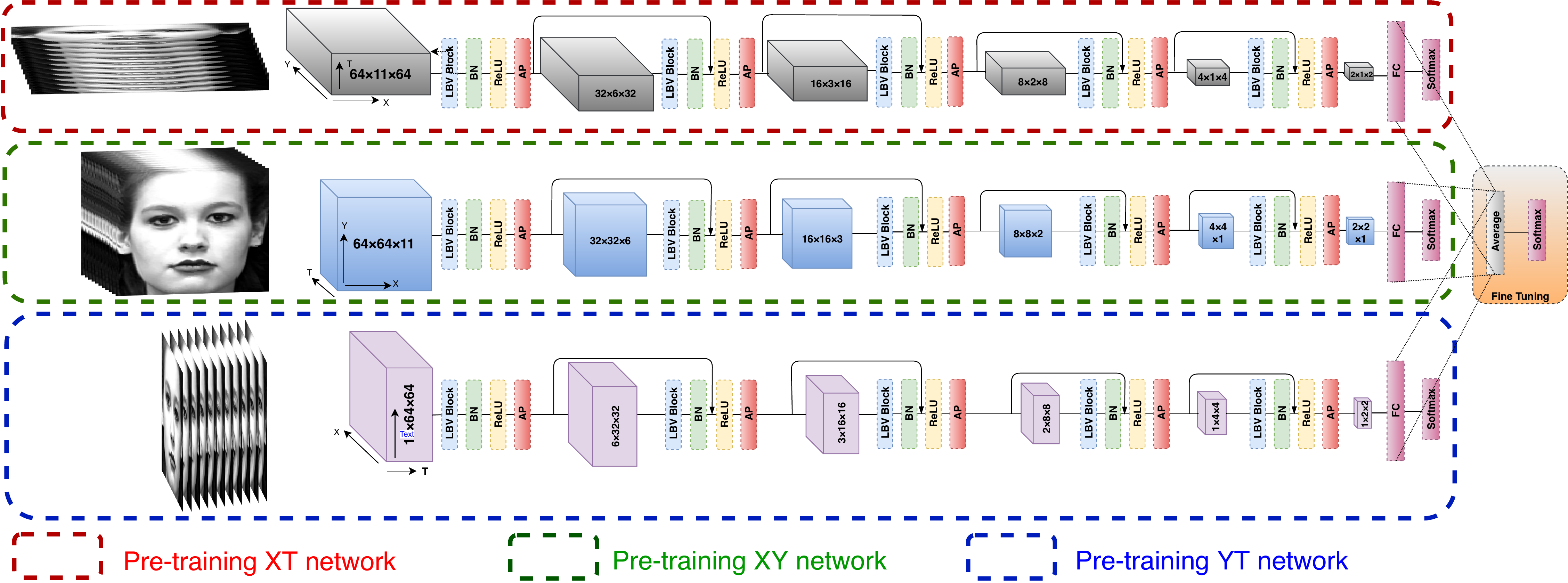}\\
	\caption{The proposed LBVCNN network architecture. BN- Batch Normalization, AP- Average Pooling, ReLU- Rectified Linear Unit. Red, green and blue dashed boxes represent individual LBVCNN-XT, LBVCNN-XY, and LBVCNN-YT networks.}
	\label{fig:block}	
\end{figure*}

\subsection{Local Binary Volume Convolutional Neural Network}\label{sec:LBVCNN}
In order to make LBCNN useful for dynamic video sequences, a straight-forward way would be to apply it in an LBP-TOP fashion. This can be done by using three separate LBCNN networks for XY, XT, and YT planes of the video cuboid and taking the third dimension as channels, and finally combining them and fine-tuning the integrated network. However, we found experimentally that such an approach fails to fully capture the spatio-temporal variations along all the dimensions. Table~\ref{tab:lbtop} shows the results when LBCNN is applied on the CK+ and Oulu-CASIA datasets in LBP-TOP fashion. Here, the evaluation is performed using 10 fold cross validation. We think that the subtle structural difference (as discussed previously in \cite{zhao2007dynamic}) between VLBP and LBP-TOP is responsible for such a phenomenon. In order to solve this problem, we propose a 3D variant of the LBC layer and integrate it with our newly proposed LBVCNN network as discussed below.\\

\begin{table}[htbp]
	\centering
	\scriptsize
	\begin{tabular}{ c | c | c }
		\hline
		& \multicolumn{2}{c}{\textbf{Accuracy (\%)}} \\
		\cline{2-3}
		\textbf{Method} & \textbf{CK+} & \textbf{Oulu-CASIA} \\
		\hline
		LBCNN-XY 	& 91.2   &  72.89\\ 
		LBCNN-XT 	& 85.65  & 69.26  \\ 
		LBCNN-YT		& 86.1 & 69.24\\ 
		\hline
		LBCNN(joint) & 92.52  & 74\\ 
		\hline  
	\end{tabular}
	\vspace{0.1in}
	\caption{Results on the CK+ and Oulu-CASIA dataset when the LBCNN network is applied in a straightforward way in the LBP-TOP fashion on the sides of video cuboid.}
	\label{tab:lbtop}
\end{table}

\noindent\textbf{Local Binary Volume Layer.} We propose a 3D variant of the LBC layer of LBCNN network that we call Local Binary Volume (LBV) layer (Fig.~\ref{fig:lbcnn}). The LBV layer is simple and very powerful in capturing subtle spatio-temporal variations in temporal image sequences. The LBV layer consists of two sub-layers. The first layer involves convolving the input with fixed non-trainable 3D filters of size $3\times3\times3$ in order to get a 3D difference map, followed by a ReLU activation to get an approximate 3D local binary bit-map. The non-trainable 3D filters contain values sampled from the set \{-1,0,1\} using Bernoulli distribution. The number of elements from the set \{-1,1\} determine the sparsity of the 3D filter. The second layer is trainable and involves $1\times1\times1$ convolutions on the output of the first layer in order to get the 3D feature maps. As proved in LBCNN \cite{juefei2017local}, we show experimentally that our LBV layer  approximates 3D convolutional layer of the conventional 3D-CNN.

Before discussing the complete structure of our proposed network, we discuss the usefulness of the ensemble of networks in deep learning and its relevance to the problem of recognizing facial expressions from videos.\\

\noindent\textbf{Usefulness of ensemble of networks.} Training and fine-tuning CNNs is difficult as it requires experimenting with many hyperparameters, and data splits and is highly subject to overfitting. An ensemble of independently trained networks can improve the predictions by reducing the overfit and can avoid the possible poor test result of a single network \cite{hansen1990neural}. However, in a data fusion ensemble model, multiple networks are necessary to analyze the heterogeneous input data \cite{ahmad2005combination}. In other words, independent networks learn different data modalities to make a collective classification decision. In general, spatial information of video for each frame is captured by XY plane, whereas the temporal variations can be observed using YT and XT planes. Fig.~\ref{fig:3dexample} shows the variations observed along all the three directional planes. Approximately only half of video cubes are shown in order to illustrate the variations clearly along all the three planes. In Fig.~\ref{fig:3dexample}, space-time visual motion impression of rows and columns can be observed using only XT and YT planes especially around eyes and lips. By combining the information from all these three planes using an ensemble of convolutional neural networks, we can extract appearance and motion information separately.\\


\noindent\textbf{Local Binary Volume Convolutional Neural Network.} Fig.~\ref{fig:block} shows the general architecture of our proposed network LBVCNN. It consists of three small 3D CNNs that we call LBVCNN-XY, LBVCNN-XT, and LBVCNN-YT which are shown as dashed lines in Fig.~\ref{fig:block}. The networks  LBVCNN-XY, LBVCNN-XT, and LBVCNN-YT capture spatio-temporal information from the three orthogonal sides of a video cuboid XY, XT, and YT respectively. We use Res-Net like structure for all the networks. More details on the input-sizes, parameters, hyperparameters, and the structures of all the networks are given in Section~\ref{sect:exp}.   \\

\noindent\textbf{Fusion fine-tuning.} For fine-tuning, we drop the final softmax layer from each of the three fully trained networks LBVCNN-XY, LBVCNN-XT, and LBVCNN-YT. Further, we combine the three networks by an element-wise average of the output of the fully connected layers which is then connected to a final softmax layer for classification.  The fine-tuning network is shown in Fig.~\ref{fig:block}. Note that during the fine-tuning process, the entire LBVCNN network (Fig.~\ref{fig:block}) is fine-tuned at a very low learning rate.\\

\noindent\textbf{Space-time complexity analysis of the LBV layer.} On comparing our proposed LBV layer with the convolutional layer (of size $3\times 3\times 3$) of the traditional 3D CNN network, we can see that the LBV layer has 27 times less trainable filters. This is due to the fact that only the second layer of LBV (with $1\times 1 \times 1$ size filters) is trainable while the first layer has fixed  non-trainable filters of size $3\times 3\times 3$.

Furthermore, the 3D convolution operation in LBV (first layer) contains just addition and subtraction operations due to the presence of -1, 0, and 1. This is in contrast to multiplicative floating point operations in a traditional 3D convolution layer.

\begin{figure*}[t]
	\centering
	\includegraphics[width=.8\textwidth]{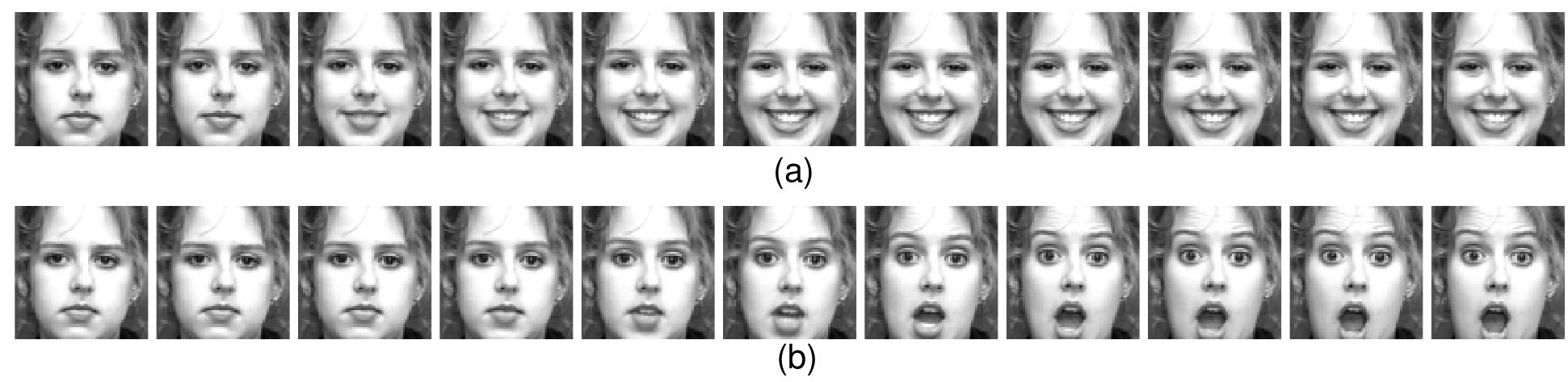}
	\caption{An example of the cropped and resized  frames from CK+ dataset of (a) happy and (b) surprise emotions.}
	\label{fig:sample}
\end{figure*}

\section{Experiments}\label{sect:exp}

To evaluate our model, we conducted extensive experiments on the three popular facial expression recognition datasets - CK+, Oulu-CASIA, and UNBC McMaster shoulder pain. We start by discussing data pre-processing and augmentation.

\noindent\textbf{Data preprocessing.} In order to process the data through the proposed network, we perform a few preprocessing steps. Note that each video consists of varying number of frames. Therefore, in order to account for varying temporal lengths, we used the video normalization method from \cite{zhou2011towards}. The method converts the video sequences of arbitrary lengths into a fixed length sequence (11 in our case). The normalized fixed length temporal patterns preserve the characteristics of the original video well \cite{zhou2011towards}. Thus, it will not affect the performance of the model. Note that another recent FER work, DTAGN \cite{jung2015joint} uses the same method for video normalization with a fixed length of 11 sequences. These 11 frames represent a neutral to peak expression. Face is extracted from each frame, cropped, and resized to 64$\times$ 64 size. Therefore, each temporal image sequence is of shape $64\times 64\times 11$ (XYT).  Sample frames of a happy and surprise expressions of a subject are shown in Fig.~\ref{fig:sample}.

\noindent\textbf{Data augmentation.} Data classification using a deep network requires a large amount of data to train the network in order to prevent overfitting. However, the datasets which have been used in this experiment contain only hundreds of videos. Hence, in order to increase the data, we perform data augmentation similar to \cite{jung2015joint}. Cropped and resized facial frames are rotated to $5^{\circ} $, $10^{\circ} $, $15^{\circ} $, $-5^{\circ} $, $-10^{\circ} $, and $-15^{\circ} $ angles. Frames are flipped and again rotated with the above six angles. Hence, a total of 14 times of original (1 original + 6 angles of original + 1 flipped + 6 angles of flipped ) dataset has been created through data augmentation. 

\noindent\textbf{Construction of the LBV layer.} As discussed in section~\ref{sec:LBVCNN}, the LBV layer consists of two convolutional layers. The first layer is a 3-D convolution layer with 64 fixed non-trainable $3\times 3\times 3$ filters. This is followed by a convolution layer containing 64 trainable $1\times 1\times 1$ filters with a ReLU activation function in between. For the first layer, we construct a filter bank of 64 - $3\times3\times3$ filters. Each of the 64 3-D filters contains values from the set \{-1,0,1\} sampled according to the Bernoulli distribution. Sparsity which is defined as the number of non-zero elements of each filter is kept as 0.9. Note that all the LBV layers in our experiments share the same 64 non-trainable 3D filters irrespective of the network (LBVCNN-XY, LBVCNN-XT, LBVCNN-YT, or joint network) that they are being used. Note that the 64 - $1\times 1\times 1$ trainable filters are not shared among the three networks and are learned independently.\\

\noindent\textbf{Network architecture.} Each of the small networks LBVCNN-XY, LBVCNN-XT, and LBVCNN-YT (see Fig.~\ref{fig:block}) take inputs of different sizes. Let XYT (in our case X=64, Y=64 and T=11) be the size of our video cuboid with XY being the spatial dimension and XT and YT being the temporal dimensions. The network LBVCNN-XY  takes as input a volume cuboid of shape $64\times 64\times 11$ while the networks LBVCNN-XT and LBVCNN-YT take inputs of sizes $(64\times 11 \times 64)$ and  $(11\times 64 \times 64)$ respectively. Rest of the network is same for all the three networks with the input layer followed by five consecutive LBV layers with a maxpooling layer after each LBV layer, except the last. The final LBV layer is followed by a fully connected layer of size 256 which is followed by a final softmax layer for classification. The architecture of the combined network for fine-tuning is discussed in Section~\ref{sec:LBVCNN}  and shown in Fig.~\ref{fig:block}. \\

\noindent\textbf{Training.} Our LBVCNN network architecture is shown in Fig.~\ref{fig:block}. At first, each of the subnetworks LBVCNN-XY, LBVCNN-XT, and LBVCNN-YT (see Fig.~\ref{fig:block}) were trained separately on XY-T, XT-Y, and YT-X cuboids respectively. All the subnetworks use adam optimizer with momentum 0.9, learning rate 1e-3, and are trained for 50 epochs. Finally, all the fully trained subnetworks are integrated as shown in Fig.~\ref{fig:block} for fine-tuning. The joint network is then fine-tuned for 100 epochs with SGD (Stochastic Gradient Descent) optimizer with momentum 0.9 and learning rate 1e-7. Throughout all our experiments, we maintain the batch-size of 16. The loss function used is categorical cross-entropy.

\noindent\textbf{Testing.} For testing, we adopt the k-fold cross-validation method. Details of the number of splits/folds created and the method used for their construction is provided in the description section of the datasets. Note that, while testing on a particular split/fold, we consider its unaugmented part only \cite{jung2015joint}.    

\subsection{CK+ Dataset}
\noindent\textbf{Description of the dataset:} Cohn-Kanade AU-Coded Expression dataset is a benchmark for facial expression recognition \cite{kanade2000comprehensive,lucey2010extended}. This dataset is composed in a restricted environment where the subject is facing the camera with an empty background. Each video in the dataset starts with a neutral expression and ends with a peak expression. Each video is labeled as an expression of anger, contempt, disgust, fear, happiness, sadness, and surprise. The dataset contains a total of 327 videos collected from 118 subjects. Each video includes a varying number of frames. For the preparation of the dataset, the subjects are arranged by ID in ascending order. These subject IDs are then partitioned into 10 subsets by sampling in ID ascending order with a step size of 10 \cite{liu2014learning}. Nine subsets were used for training and the remaining one was used for validation \cite{liu2014learning}. This process is called as 10-fold cross-validation. The evaluation is performed in a subject independent way.\\

\begin{figure}[t]
	\centering
	\includegraphics[height=0.5\columnwidth, width=\columnwidth]{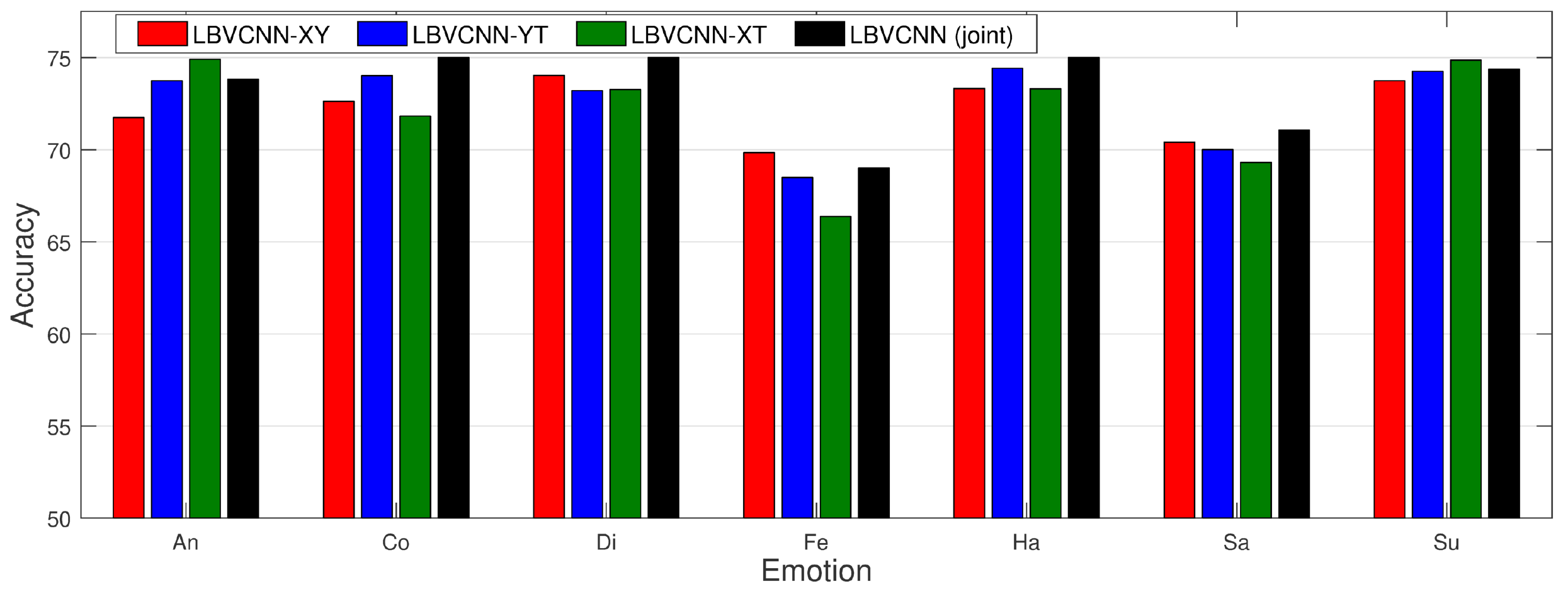}
	\caption{Comparison of accuracy according to each emotion among four networks on Oulu-CASIA dataset.}
	\label{fig:ck}
\end{figure}

\noindent\textbf{Results:} The total accuracy  of 10-fold cross-validation of our model on the CK+ dataset is shown in Table~\ref{tab:ck_all}. Note that in order to make the comparison fair, we do not consider image-based and 3D geometry based algorithms and models from the comparison tables. The top three models DTAGN \cite{jung2015joint}, LOMo \cite{sikka2016lomo} and PHRNN-MSCNN \cite{zhang2017facial} that have recently achieved state-of-the-art accuracy use facial landmarks. Our model achieves state-of-the-art accuracy when compared to the models like HOG 3D \cite{klaser2008spatio}, Cov3D \cite{sanin2013spatio}, and STM-ExpLet \cite{liu2014learning} that do not use facial landmarks. It achieves results comparable to landmark-based state-of-the-art models and better results when compared to the non landmark-based models.
\begin{table}[htbp]
	\centering
	\scriptsize
	\begin{tabular}{| c | c | c | c |}
		\hline
		\textbf{Method} & \textbf{Accuracy}  & \textbf{Landmarks} & \textbf{Strategy} \\
		\hline
		HOG 3D \cite{klaser2008spatio}  	& 91.44	& $\times$		 & 10 folds \\
		TMS \cite{jain2011facial} 			& 91.89 & $\checkmark$   & 4 folds \\
		Cov3D \cite{sanin2013spatio} 		& 92.30 & $\times$       & 5 folds\\
		3DCNN-DAP \cite{liu2014deeply} 		& 92.40 & $\checkmark$   & 15 folds \\
		STM-ExpLet \cite{liu2014learning} 	& 94.19 & $\times$       & 10 folds\\
		LOMo \cite{sikka2016lomo}			& 95.10 & $\checkmark$	 & 10 folds\\
		VLBP \cite{zhao2007dynamic}  		& 96.26	& $\times$		 & 10 folds \\
		DTAGN \cite{jung2015joint} 			& 97.25 & $\checkmark$   & 10 folds \\
		PHRNN-MSCNN\cite{zhang2017facial}   & 98.50 & $\checkmark$   & 10 folds\\
		\hline
		LBVCNN-XY 							& 95.31 & $\times$ 		 & 10 folds \\ 
		LBVCNN-XT 							& 95.50 & $\times$ 		 & 10 folds  \\ 
		LBVCNN-YT 							& 95.19 & $\times$       & 10 folds \\ 
		LBVCNN(joint) 						& 97.38 & $\times$       & 10 folds \\ 	
		\hline 
	\end{tabular}
	\vspace{0.1in}
	\caption{Comparison of various methods on the CK+ dataset in terms of average recognition accuracy of seven expressions. Note that in order to make the comparison fair, we do not consider image-based and 3D geometry based algorithms and models. VLBP \cite{zhao2007dynamic} results are for six expressions only.}
	\label{tab:ck_all}
\end{table}
\begin{table}[htbp]
	\centering
	\scriptsize
	\begin{tabular}{| c | c | c | c | c | c| c | c | }
		\hline
		& \textbf{An} & \textbf{Co} & \textbf{Di} & \textbf{Fe} & \textbf{Ha} & \textbf{Sa} & \textbf{Su} \\
		\hline
		\textbf{An} & 97.63 & 0 & 2.37 & 0 & 0 & 0 & 0\\
		\hline
		\textbf{Co} & 0 & 100 & 0 & 0 & 0 & 0 & 0 \\
		\hline
		\textbf{Di} & 0  & 0 & 100 & 0 & 0 & 0 & 0\\
		\hline
		\textbf{Fe} & 0  & 0 & 0 & 88.05 & 7.97 & 3.98 &0 \\
		\hline
		\textbf{Ha} & 0 & 0 & 0 & 0 & 100 & 0 & 0 \\
		\hline
		\textbf{Sa} & 2.62 & 2.62 & 0 & 2.62 & 0 & 92.14 & 0  \\
		\hline
		\textbf{Su} & 0 & 0 & 0 & 0 & 1.28 & 0 & 98.72 \\
		\hline	    
	\end{tabular}
	\vspace{0.1in}
	\caption{Confusion matrix of LBVCNN (joint) on CK+ dataset.}
	\label{tab:ck}
\end{table}
\begin{figure}[htbp]
	\centering
	\includegraphics[height=0.5\columnwidth, width=\columnwidth]{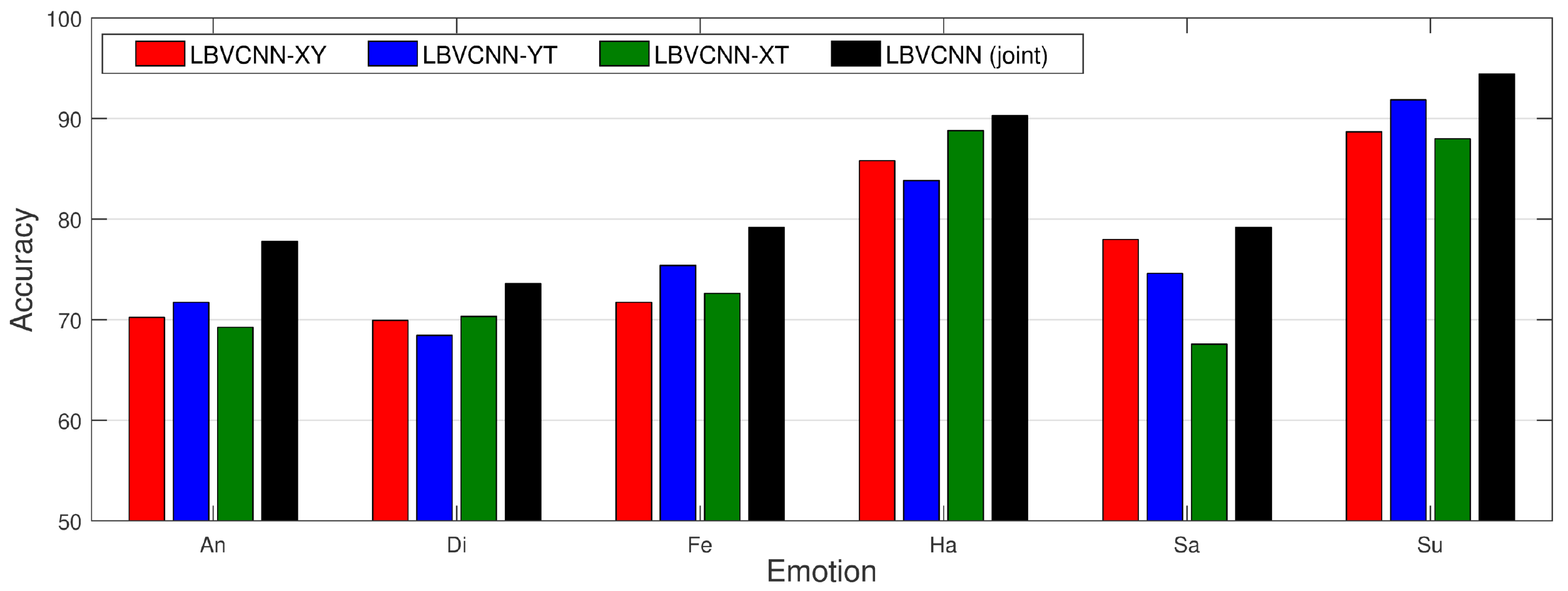}
	\caption{Comparison of accuracy according to each emotion among four networks on CK+ dataset.}
	\label{fig:oulu}
\end{figure}
The confusion matrix of the combined network i.e., LBVCNN (joint) on CK+ dataset is reported in Table ~\ref{tab:ck}. Comparison of accuracy according to each emotion among four networks is shown in Fig.~\ref{fig:ck}. The accuracy in the cases of angry, contempt, disgust, happiness, and surprise is good, but the performance for sadness and fear is relatively poor. 

\subsection{Oulu-CASIA Dataset}
\noindent\textbf{Description of the dataset:} The Oulu-CASIA dataset consists of six expressions (surprise, happiness, sadness, anger, fear, and disgust) from 80 subjects under visible light condition \cite{zhao2011facial}. Subjects are between 23 to 58 years old and 73.8$\% $ of the subjects are males. It has a total of 480 video sequences, 6 each for 80 subjects. The dataset provides cropped version (only face) of the original frames. We have performed similar preprocessing as in CK+ dataset on cropped frames. Only 11 frames per video are considered and resized to 64 $\times $ 64. Each video sequence starts with a neutral expression and ends with a peak expression. Preparation of this dataset is done similarly to that of CK+ dataset.\\

\noindent\textbf{Results:} The total accuracy  of 10-fold cross-validation of our model on the Oulu-CASIA dataset is shown in Table~\ref{tab:oulu_all}. Note that the models DTAGN \cite{jung2015joint}, LOMo \cite{sikka2016lomo}, and PHRNN-MSCNN \cite{zhang2017facial} that have recently achieved state-of-the-art accuracy use facial landmarks. Thus, geometric features certainly boost the performance of expression recognition models. Our model achieves state-of-the-art accuracy when compared to the models like HOG 3D \cite{klaser2008spatio}, AdaLBP  \cite{zhao2011facial}, and STM-ExpLet \cite{liu2014learning} that do not use facial landmarks. It achieves results comparable to the landmark-based state-of-the-art models except the PHRNN-MSCNN\cite{zhang2017facial} and better results when compared to all the non landmark-based models.
The confusion matrix of the combined network i.e. LBVCNN (joint) is reported in Table ~\ref{tab:oulu}. Comparison of accuracy according to each emotion among the four networks is shown in Fig.~\ref{fig:oulu}. The accuracy in the cases of fear, happiness, sadness, and surprise is good, but the performance for anger and disgust is relatively poor. In particular, there is a high degree of confusion among the expressions anger, disgust and sadness.

\begin{table}[t]
	\centering
	\scriptsize
	\begin{tabular}{| c | c | c | c | }
		\hline
		\textbf{Method} & \textbf{Accuracy} & \textbf{Landmarks} & \textbf{Strategy}\\
		\hline
		HOG 3D 	\cite{klaser2008spatio} 	& 70.63 & $\times$     & 10 folds\\
		AdaLBP  \cite{zhao2011facial}		& 73.54 & $\times$	   & 10 folds\\
		STM-ExpLet \cite{liu2014learning}	& 74.59 & $\times$ 	   & 10 folds\\
		DTAGN \cite{jung2015joint} 			& 81.46 & $\checkmark$ & 10 folds\\	
		LOMo \cite{sikka2016lomo}			& 82.10 & $\checkmark$ & 10 folds\\	
		PHRNN-MSCNN\cite{zhang2017facial}   & 86.25 & $\checkmark$ & 10 folds\\
		\hline
		LBVCNN-XY 							& 77.40 & $\times$ & 10 folds \\ 
		LBVCNN-XT 							& 77.59 & $\times$ & 10 folds \\ 
		LBVCNN-YT						    & 76.09 & $\times$ & 10 folds \\ 
		LBVCNN(joint) 						& 82.41 & $\times$ & 10 folds\\ 
		\hline  
	\end{tabular}\vspace{.1in}
	\caption{Comparison of various methods on the Oulu-CASIA dataset in terms of average recognition accuracy of six expressions.  Note that in order to make the comparison fair, only video based methods are included.}	
	\label{tab:oulu_all}
\end{table}
\begin{table}[t]
	\centering
	\scriptsize
	\begin{tabular}{| c | c | c | c | c | c| c | }
		\hline 
		& \textbf{An}  & \textbf{Di} & \textbf{Fe} & \textbf{Ha} & \textbf{Sa} & \textbf{Su} \\
		\hline
		\textbf{An} & 77.78  & 6.94  & 4.17 &  0  & 11.11  & 0  \\
		\hline
		\textbf{Di} & 13.89   & 73.61 & 1.39  & 2.78 & 8.33  &  0 \\
		\hline
		\textbf{Fe} & 0  & 5.56 & 79.17 & 2.78 & 5.56  & 6.94  \\
		\hline
		\textbf{Ha} & 1.39 & 1.39 & 5.56 & 90.28 & 1.39  & 0 \\
		\hline
		\textbf{Sa} & 12.5  & 5.56 & 2.78 & 0 & 79.17  & 0   \\
		\hline
		\textbf{Su} & 0 & 1.39 & 4.17 & 0 & 0  & 94.44  \\
		\hline
	\end{tabular}
	\vspace{0.1in}
	\caption{Confusion matrix of LBVCNN (joint) on Oulu-CASIA dataset.}
	\label{tab:oulu}
\end{table}


\subsection{UNBC McMaster Shoulder Pain Dataset}
\noindent\textbf{Description of the dataset:} Unlike CK+ and Oulu-CASIA datasets which are in controlled setting, UNBC McMaster dataset  is in spontaneous setting \cite{lucey2011painful}. This makes the task of facial expression recognition even more challenging. The dataset consists of real
world videos of subjects with pain while performing guided movements of their affected and unaffected arms in a clinical interview. The videos are rated for pain intensity (0 to 5) by trained experts. Following \cite{sikka2016lomo}, we labeled videos as ``pain" for intensity above 3 and ``no pain" for intensity 0, and discarded the rest.  This resulted in 149 videos from 25 subjects with 57 positive and 92 negative samples. Following \cite{ruiz2014regularized}, a temporal window of 0.5 seconds is taken. The process of data pre-processiong and augmentation is same as that of CK+ and the Oulu-CASIA datasets. Unlike the case of CK+ and Oulu-CASIA datasets, the validation protocol used is ``leave one subject out" which is same as the works mentioned in Table~\ref{tab:unbc}.\\

\noindent\textbf{Results:}  The total accuracy  of ``leave one subject out" cross-validation of our model on the UNBC McMaster dataset is shown in Table~\ref{tab:unbc}. Note that the models MS-MIL \cite{sikka2014classification}, MIL-HMM \cite{wu2015multi}, RMC-MIL \cite{ruiz2014regularized}, and LOMo \cite{sikka2016lomo} use landmarks to achieve state-of-the-art results. Our method achieves better results than all, except the LOMo \cite{sikka2016lomo} where we achieve comparable results.

\begin{table}[htbp]
	\centering
	\scriptsize
	\begin{tabular}{| c | c | c | }
		\hline
		\textbf{Method} & \textbf{Accuracy} & \textbf{Landmarks} \\
		\hline
		MS-MIL \cite{sikka2014classification} 			& 83.7 & $\checkmark$ \\
		MIL-HMM\cite{wu2015multi}   & 85.2 & $\checkmark$ \\
		RMC-MIL\cite{ruiz2014regularized}   & 85.7 & $\checkmark$ \\	
		LOMo\cite{sikka2016lomo}			& 87.0 & $\checkmark$ \\	
		\hline
		LBVCNN-XY 							& 84.76 & $\times$  \\ 
		LBVCNN-XT 							& 83.20 & $\times$  \\ 
		LBVCNN-YT						    & 83.48 & $\times$  \\ 
		LBVCNN(joint) 						& 86.55 & $\times$ \\ 
		\hline  
	\end{tabular}\vspace{.1in}
	\caption{Comparison of various methods on the UNBC McMaster shoulder pain dataset in terms of average recognition accuracy of pain and no pain expressions.}	
	\label{tab:unbc}
\end{table}

\subsection{Feature Visualization}

\begin{figure}[t]
	\includegraphics[width=.37\textwidth]{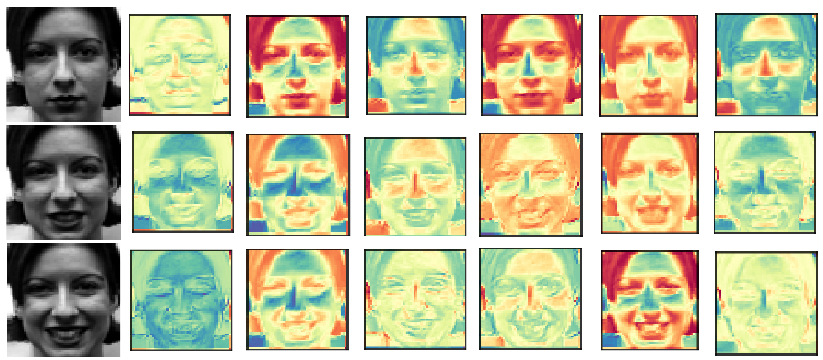} 
	\includegraphics[width=.37\textwidth]{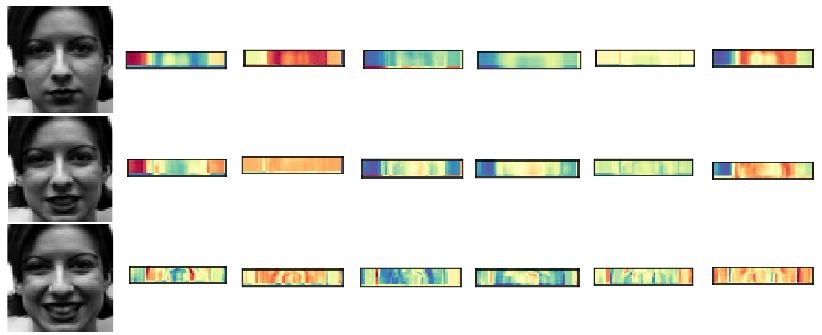} 
	\includegraphics[width=.33\textwidth]{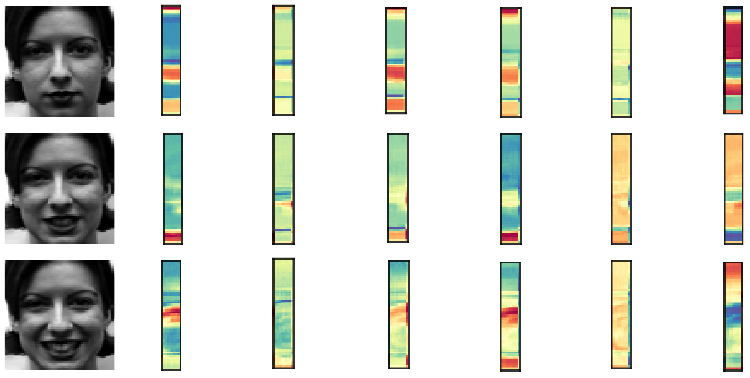} 
	\caption{Feature maps learned by the LBVCNN-XY (left), LBVCNN-XT (middle), and LBVCNN (right) for the happy emotion. Blue and red represent the  high and low response values.}	
	\label{fig:happy}
\end{figure}

\begin{figure}[htbp]
	\centering
	\includegraphics[width=.52\columnwidth]{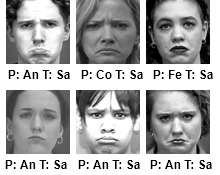} 
	\includegraphics[width=.42\columnwidth]{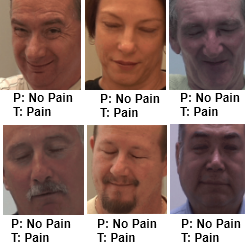} 
	\caption{Failure cases from the CK+ (left) and UNBC (right) datasets. P- Predicted, T- Target.}
	\label{fig:fail}
\end{figure}

In this section, we visualize the learned feature maps  of our LBVCNN model. Fig.~\ref{fig:happy} show  the feature maps learned by our multi frame-based CNN in the first layer for expressions angry, happy, and surprise respectively on the CK+ database. For the sake of simplicity, only six out of sixty four filters feature maps are shown. Here, blue and red represent the  high and the low response values. We observe that our model is able to capture the facial expression movements very effectively. Furthermore, the learned feature maps are consistent, for example the feature maps corresponding to the starting frame which is a neutral frame for each emotion sequence have approximately same visualization. Fig.~\ref{fig:fail} show the failure cases from CK+ and UNBC McMaster shoulder pain datasets, respectively. We observe that, for UNBC McMaster dataset the videos with true label as ``pain" and misclassified as ``no pain" are high. The number of cases where videos with true label as ``no pain" being misclassified as ``pain" are very less.



\section{Conclusion}\label{sect:conclusion}
A novel 3D-CNN is proposed in order to recognize facial expressions from image sequences in an end-to-end fashion. The method can be performed directly on image sequences without any additional information such as facial landmarks. In particular, local binary volume layer (an efficient replacement of 3D-CNN layer) is proposed based on the concept of volume local binary pattern. LBV layer saves a significant number of trainable parameters when compared to conventional 3D-CNN layer. Our proposed network, LBVCNN, achieves comparable results on CK+, Oulu-CASIA and UNBC McMaster shoulder pain datasets. Most of the state-of-art methods use facial landmarks to extract geometric features. Since detecting landmarks is a difficult problem by itself and the problem becomes more complex with changes in illumination, resolution, and orientation, our work is of significant use as it does not use landmarks to drive the expression recognition process.


In future, we will seek utilization of local binary volume layer in other face video based computer vision problems such as face recognition and biometrics (e.g., age, ethnicity, gender recognition). In such problems, geometric features (e.g. facial landmarks) are used to boost the accuracy of the models. We shall explore other video based applications where additional features are required to boost the accuracy of the model.

\noindent\textbf{Acknowledgments.} Sudhakar Kumawat was supported by TCS Research Fellowship. Shanmuganathan Raman was supported by SERB Core Research Grant and Imprint 2 Grant.

\end{document}